\documentclass{article}

\usepackage{PRIMEarxiv}

\usepackage[utf8]{inputenc} 
\usepackage[T1]{fontenc}    
\usepackage{hyperref}       
\usepackage{url}            
\usepackage{booktabs}       
\usepackage{amsfonts}       
\usepackage{nicefrac}       
\usepackage{microtype}      
\usepackage{lipsum}
\usepackage{fancyhdr}       
\usepackage{graphicx}       
\graphicspath{{media/}}     

\title{Personalized Prediction of Offensive News Comments \\by Considering the Characteristics of Commenters}

\author{
  Teruki Nakahara\\
  Kyushu University \\
  \texttt{nakahara.teruki.528@s.kyushu-u.ac.jp} \\
   \And
  Taketoshi Ushiama \\
  Kyushu University \\
  \texttt{ushiama@design.kyushu-u.ac.jp} \\
}

\begin{document}
\maketitle

\begin{abstract}
When reading news articles on social networking services and news sites, readers can view comments marked by other people on these articles. By reading these comments, a reader can understand the public opinion about the news, and it is often helpful to grasp the overall picture of the news. However, these comments often contain offensive language that readers do not prefer to read. This study aims to predict such offensive comments to improve the quality of the experience of the reader while reading comments. By considering the diversity of the readers' values, the proposed method predicts offensive news comments for each reader based on the feedback from a small number of news comments that the reader rated as ``offensive’’ in the past. In addition, we used a machine learning model that considers the characteristics of the commenters to make predictions, independent of the words and topics in news comments. The experimental results of the proposed method show that prediction can be personalized even when the amount of readers' feedback data used in the prediction is limited. In particular, the proposed method, which considers the commenters' characteristics, has a low probability of false detection of offensive comments.
\end{abstract}


\section{Introduction}
\label{Introduction}
In recent years, reading the news on the Internet has become a common practice. According to The Japan Press Research Institute\cite{shinbun2021}, 73.1\% of respondents from all age groups read the news on the Internet at least once a week, and among those in their 40s or less, the percentage increases to 90\%. In addition to portals and news sites operated by TV stations and newspapers, many people read the news on social networking services (SNSs). In particular, young people in their teens and twenties most frequently use SNSs to access news through the Internet\cite{shinbun2021}.

When reading the news on the Internet, readers can read not only the news but also other people's comments on the news. Yahoo News\footnote{https://news.yahoo.co.jp/}, one of the leading news sites in Japan, has a comments section for each news article. On Twitter\footnote{https://twitter.com/}, one of the most popular SNSs, social-media accounts of TV stations and newspapers post news and links to news articles in the form of tweets. Users can post comments on news stories as replies to these tweets, similar to how they post comments on news sites.

In a survey by Stroud et al.\cite{stroud2016news}, 49\% of the people who read news on the Internet said that they also read comments. By reading other people's comments, they can understand the public opinion about the news, obtain an overall picture, and understand it better. However, in online communities, such as news comment sections, readers may find offensive comments, such as those containing extreme or slanderous language.

To solve this problem, methods for predicting comments that are offensive to readers have been proposed\cite{10.1145/3200947.3208069, perspective,hessel-lee-2019-somethings}. These methods use machine learning models that predict whether a comment is offensive to readers by inputting the comment's text and other information. However, the tendency of comments that readers find offensive differs depending on each reader's values and other factors\cite{sap-etal-2022-annotators}.

\begin{figure}[tb]
\begin{center}
\includegraphics[width=120mm]{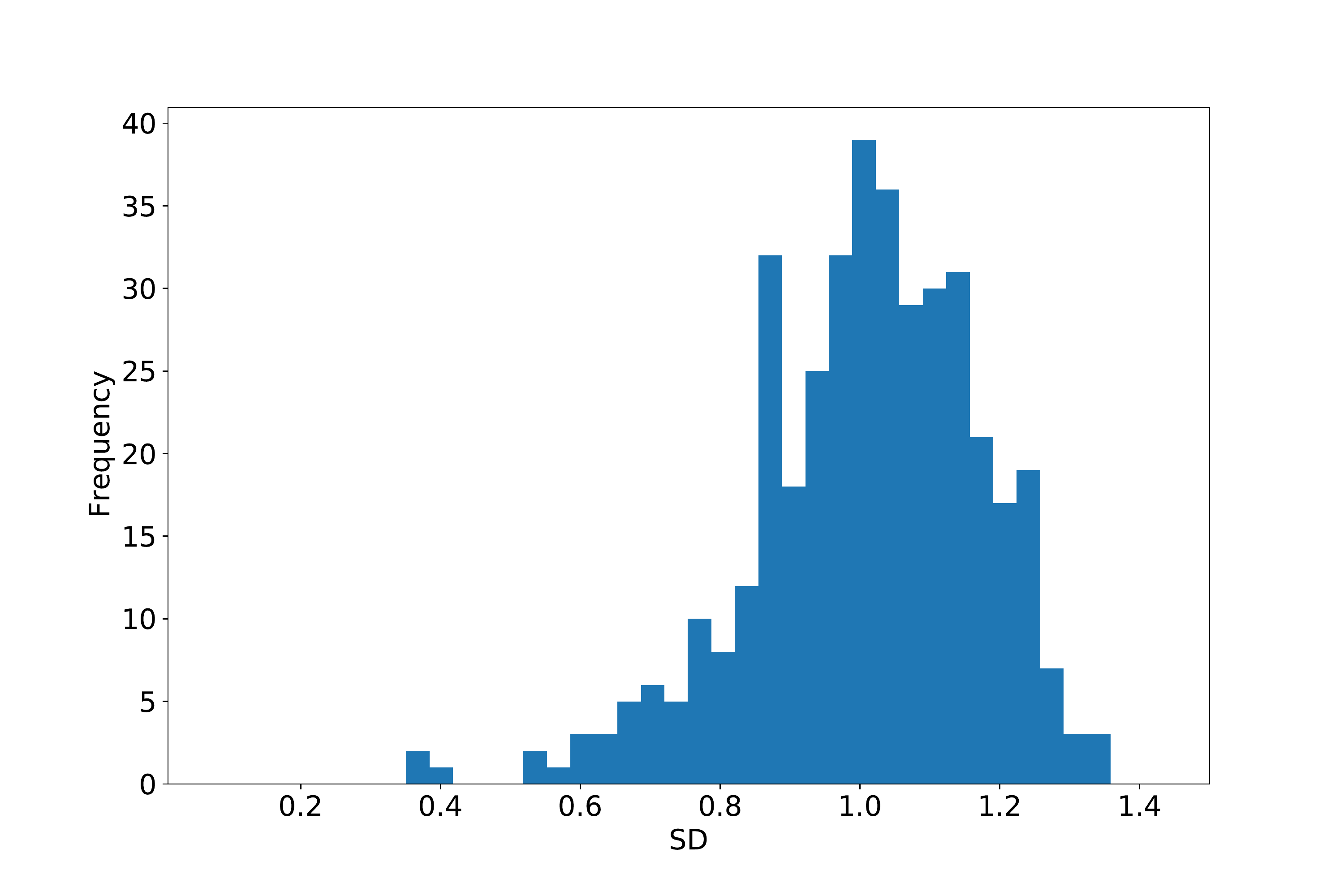}
\end{center}
\caption{Distribution of standard deviations in evaluation values}
\label{fig:Distribution of standard deviations in evaluation values}
\end{figure}

To test the importance of personalization in predicting offensive news comments, we examined the variation in ratings of comments among comment readers. We conducted a survey on the crowdsourcing service CrowdWorks\footnote{https://crowdworks.jp/} and labeled the news comments. In the survey, 400 news comments posted during the three months from April 1 to June 30, 2022, were presented to 50 users. They were asked to rate the comments on a five-point scale regarding whether they found them offensive \{1: Strongly disagree, 2: Disagree, 3: Neither agree nor disagree, 4: Agree, 5: Strongly agree\}. In this way, we obtained responses from 50 users per news comment and calculated the standard deviation of the evaluation values for each comment. Examples of the relationship between comments and standard deviations are listed in Table \ref{tab:Example of the relationship between comments and standard deviation}. A histogram of the standard deviations of the 400 comments is shown in Figure \ref{fig:Distribution of standard deviations in evaluation values}. Figure \ref{fig:Distribution of standard deviations in evaluation values} shows that the standard deviations of the five-point scale are concentrated around 1.0, indicating a variation in the evaluation of comments among  comment readers. Therefore, it is essential to personalize the prediction of offensive news comments to consider the values of comment readers.

\begin{table*}[tb]
\caption{Example of the relationship between comments and standard deviation}
\label{tab:Example of the relationship between comments and standard deviation}
\centering
\begin{tabular}{c|p{20em}|p{20em}}
\hline\hline
SD & News & Comments\\ \hline
0.58 & ``Internet Explorer’’ support ends today. & I miss...\\
0.95 & More than 100 subway stations are designated as ballistic missile evacuation facilities for the first time. & Stupid woman. With the current constitution, we'd be destroyed before we could evacuate!\\
1.26 & Prime Minister Kishida ``Improve food self-sufficiency and strengthen the international competitiveness of agriculture.’’  & Since the LDP ruined this, the party should be dissolved first.\\
\hline 
\end{tabular}
\end{table*}

Personalization methods for predicting offensive comments have been proposed and proven to improve prediction accuracy\cite{274443}. Existing methods personalize the prediction of offensive comments by using existing feedback from readers who have previously rated comments as ``offensive.’’

This study aims to generate predictions that are independent of the words and topics in the feedback. Specifically, offensive news comments are predicted by generating embeddings of readers using a machine learning model based on the characteristics of users who posted comments that readers rated as ``offensive.’’ In addition, this study used data collected from news tweets by the news TV stations' Twitter accounts and the replies (i.e., comments) to these tweets, and correct answer labels were assigned using a questionnaire.

The contributions of this study can be summarized as follows:
\begin{itemize}
    \item We investigated the variation in the ratings of comments among comment readers and proved the importance of personalization in predicting offensive news comments.
    \item We personalized the prediction of offensive news comments by generating embeddings of readers based on a small amount of feedback from readers of news comments.
    \item To generate embeddings of readers, we utilized comments that readers have rated as ``offensive’’ in the past and embeddings of users who have posted comments rated as ``offensive’’ by readers.
    \item We evaluated the performance of the proposed method by personalizing the prediction of offensive news comments using a dataset that comprises news and comments that have been posted.
 \end{itemize}
 
The remainder of this paper is organized as follows. Section 2 discusses related work. Section 3 describes the personalization of offensive news comment predictions addressed in this study. Section 4 describes the proposed method. Section 5 describes the method for generating embeddings of commenters used in the proposed method. Section 6 describes and reports the results of our experiments. Finally, Section 7 summarizes the paper and discusses future research directions.

\section{RELATED WORK}
\label{RELATED WORK}

\subsection{Support for reading news comments}
\label{Support for reading news comments}
Many studies have been conducted to improve the experience of reading news comments. For example, Ma et al.\cite{10.1145/2396761.2396798} clustered news comments by estimating the topics of comments using latent Dirichlet allocation (LDA)\cite{10.5555/944919.944937}, a topic model. Aker et al.\cite{10.1007/978-3-319-30671-1_2} performed clustering of news comments by modeling the similarity among comments using a graph-based method. In these studies, clustering assists readers in viewing news comments by providing them with an overall picture of the news comments. In this study, we improve the quality of the reader experience and promote news comment reading by predicting comments that readers may find offensive, such as those containing extreme content or slanderous remarks.

\subsection{Offensive Comments Prediction}
\label{Offensive Comments Prediction}
Many studies and implementations have been conducted to predict comments that readers may find offensive, such as toxic or insulting comments, which are posted in online communities, such as SNSs. For example, Georgakopoulos et al.\cite{10.1145/3200947.3208069} used a convolutional neural network (CNN)-based machine learning model to predict toxic comments on editorial discussion pages on Wikipedia. Hessel et al.\cite{hessel-lee-2019-somethings} classified controversial and non-controversial posts on Reddit using models, such as LSTM\cite{6795963} and BERT\cite{devlin-etal-2019-bert}. Saveski et al.\cite{10.1145/3442381.3449861} analyzed the relationship between conversational structure and the toxicity of posts on Twitter. Google Jigsaw provides a system for detecting toxic or insulting comments in Perspective API\cite{perspective}.

\subsection{Personalization of Offensive Comment Prediction}
\label{Personalization of Offensive Comments Prediction}
Sap et al.\cite{sap-etal-2022-annotators} described a relationship between the people's identity and beliefs and their toxicity rating in evaluating the toxicity comments. Therefore, it is essential to personalize the prediction of offensive comments by considering differences in the values of comment readers.

Kumar et al.\cite{274443} proposed that prediction accuracy can be improved by tuning the parameters of Perspective API provided by Google Jigsaw for each reader, using the feedback of comments that readers have rated as offensive in the past. Generic predictions of comments for personalization can be generated using such feedback, rather than collecting data from specific SNSs or news sites, such as readers' profiles or past postings. However, providing a large amount of feedback is burdensome for readers. Therefore, it is desirable to use only a small amount of feedback data in the prediction.

\section{Personalization of Offensive News Comment Prediction}
\label{Personalization of Offensive News Comments Prediction}
This study addresses the problem of predicting whether a reader of news comments will find the new news comments offensive. For personalization, a general-purpose prediction method is proposed that uses only the feedback of comments rated as ``offensive’’, without using existing data on specific SNSs or news sites, such as readers' profiles or past postings.

We assume that we are given a set of news tweets $NT=\{nt_{1},\ldots, nt_{|NT|}\}$, a set of comments $C=\{c_{1},\ldots,c_{|C|}\}$, a commenters set $U=\{u_{1},\ldots,u_{|U|}\}$, and a readers set $R=\{r_{1},\ldots,r_{|R|}\}$. 
Let $\mathrm{news}(c)$ denote the news tweet to which the comment $c$ refers and $\mathit{commenter}(c)$ denote the commenter who writes the comment $c$.
Here, $\mathit{news}(c) \in NT$ and $\mathit{commenter}(c)\in U$. 
Let $P(o \mid c,r)$ denote the probability that the user $r$ judges the comment $c$ to be offensive.
In this study, we propose a method to predict this probability using machine learning models.
The proposed method considers that the set of news comments rated as ``offensive’’ by reader $r$ is $\mathit{offensive}(r)$. 
The study aims to predict whether reader $r$ will find a new news comment $c$ offensive.




\subsection{Simple Offensive News Comment Prediction Method}
\label{Simple Offensive News Comments Prediction Method}

\begin{figure}[tb]
\begin{center}
\includegraphics[width=100mm]{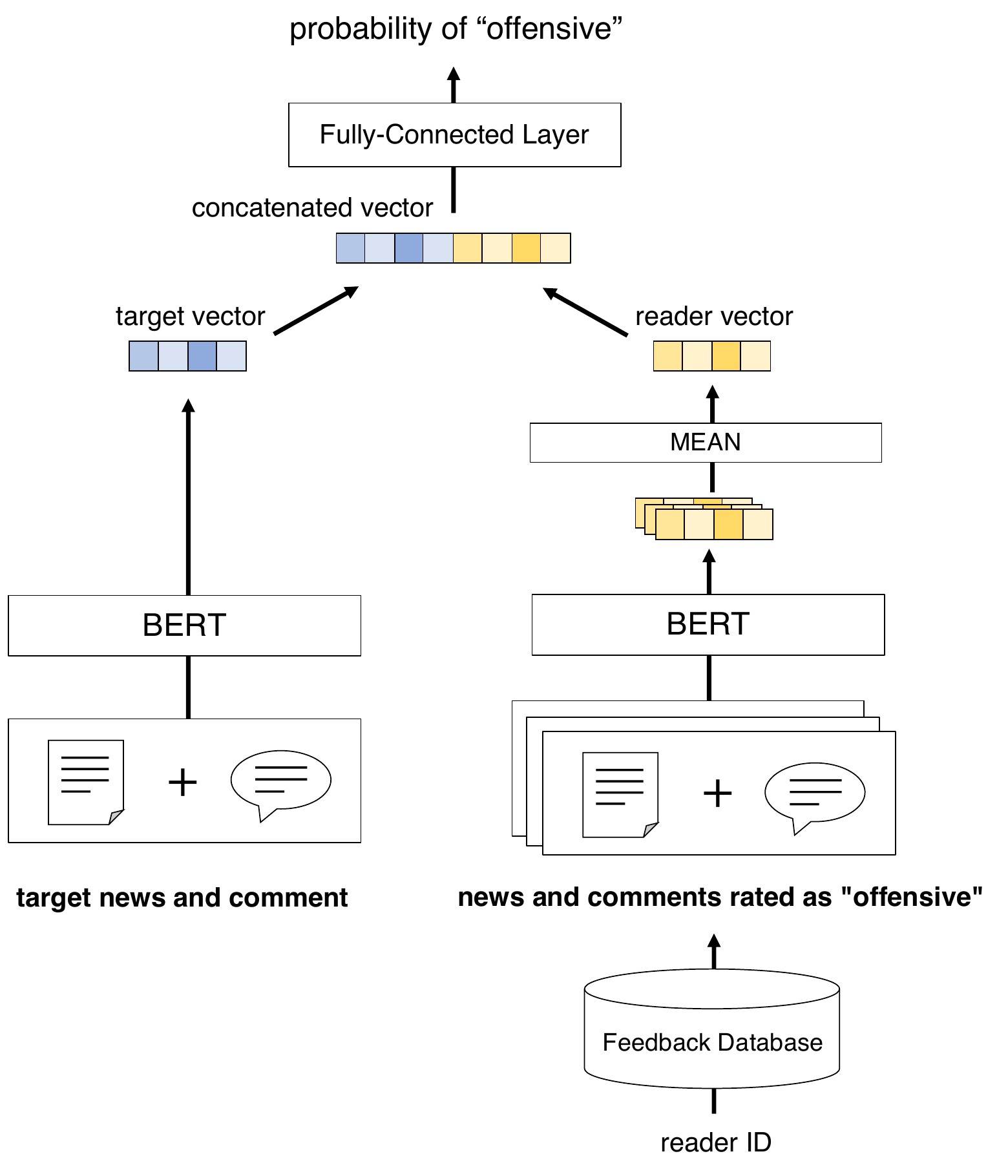}
\end{center}
\caption{Simple prediction model for offensive news comments}
\label{fig:Simple prediction model for offensive news comments}
\end{figure}

In this section, we describe a simple prediction method that serves as a baseline machine learning model for personalizing the prediction of offensive news comments. 
An overview of the simple prediction model is shown in Figure \ref{fig:Simple prediction model for offensive news comments}. This model outputs the probability that a given news comment belongs to the ``offensive’’ label based on the vector that concatenates the target and reader vectors. Therefore, the simple prediction model accepts two types of data as input. The first input is the pair of news and comments to be predicted. We vectorize the concatenated news and comment text using BERT, a natural language processing model. 
The second input is reader ID. 
In this study, we constructed a Feedback Database, which stores news comments rated as ``offensive’’ by readers alongside those readers' IDs. 
In the simple prediction model, based on the reader IDs, we retrieve several pairs of news and comment texts from the Feedback database that have been rated as ``offensive’’ in the past.
Each output vector obtained by inputting concatenated news and comment texts into BERT is averaged to generate a reader vector of comments.

The aforementioned approach can be formalized as follows:
The following equation represents a simple machine learning model that predicts the probability that user $r$ will find comment $c$ offensive:
\begin{equation}
y=f^{sim}(x;\theta^{sim}),
\end{equation}
where $x$ is a vector representation of the pair of reader $r$ and comment $c$, $y$ is a scalar value representing the probability value, and $\theta^{sim}$ is the set of parameters.
The training data set $D^{sim}$ for the simple model is defined as the set of pairs $(x,y)$ of input $x$ and output $y$ as follows:
\begin{eqnarray}
D^{sim}=\{(x,y)\mid x=em^{sim}_c(c)\oplus em^{sim}_r(r), c\in C, r\in R, y\in \{1,0\} \}
\end{eqnarray}
where $x$ is the concatenation of the embedding vector of comment $c$ and the embedding vector of reader $r$.
Furthermore, $em^{sim}_c(c)$ and $em^{sim}_r(r)$ represent the vectorization of comment $c$ and reader $r$, respectively, and $\oplus$ represents the vector concatenation operation.

By inputting the text of comment $c$ and the news tweet $news(c)$ it refers into BERT, the embedded vectorization function $em^{sim}_c(c)$ for comment $c$ is obtained and is defined as follows:
\begin{eqnarray}
em^{sim}_c(c) = \mathrm{BERT}(c,news(c))
\end{eqnarray}
By contrast, the embedded vectorization function $em^{sim}_r(r)$ of a reader $r$ is defined as follows, using the set $\mathit{offensive}(r)$ of comments deemed offensive by the reader.
\begin{eqnarray}
em^{sim}_r(r) = \sum_{x \in \mathit{offensive}(r)}{\frac{\mathrm{BERT}(x, \mathit{news}(x))}{\mid \mathit{offensive}(r)\mid}}
\end{eqnarray}

\subsection{BERT for Vectorizing News and Comments}
\label{BERT for Vectorizing News and Comments}

\begin{figure}[tb]
\begin{center}
\includegraphics[width=100mm]{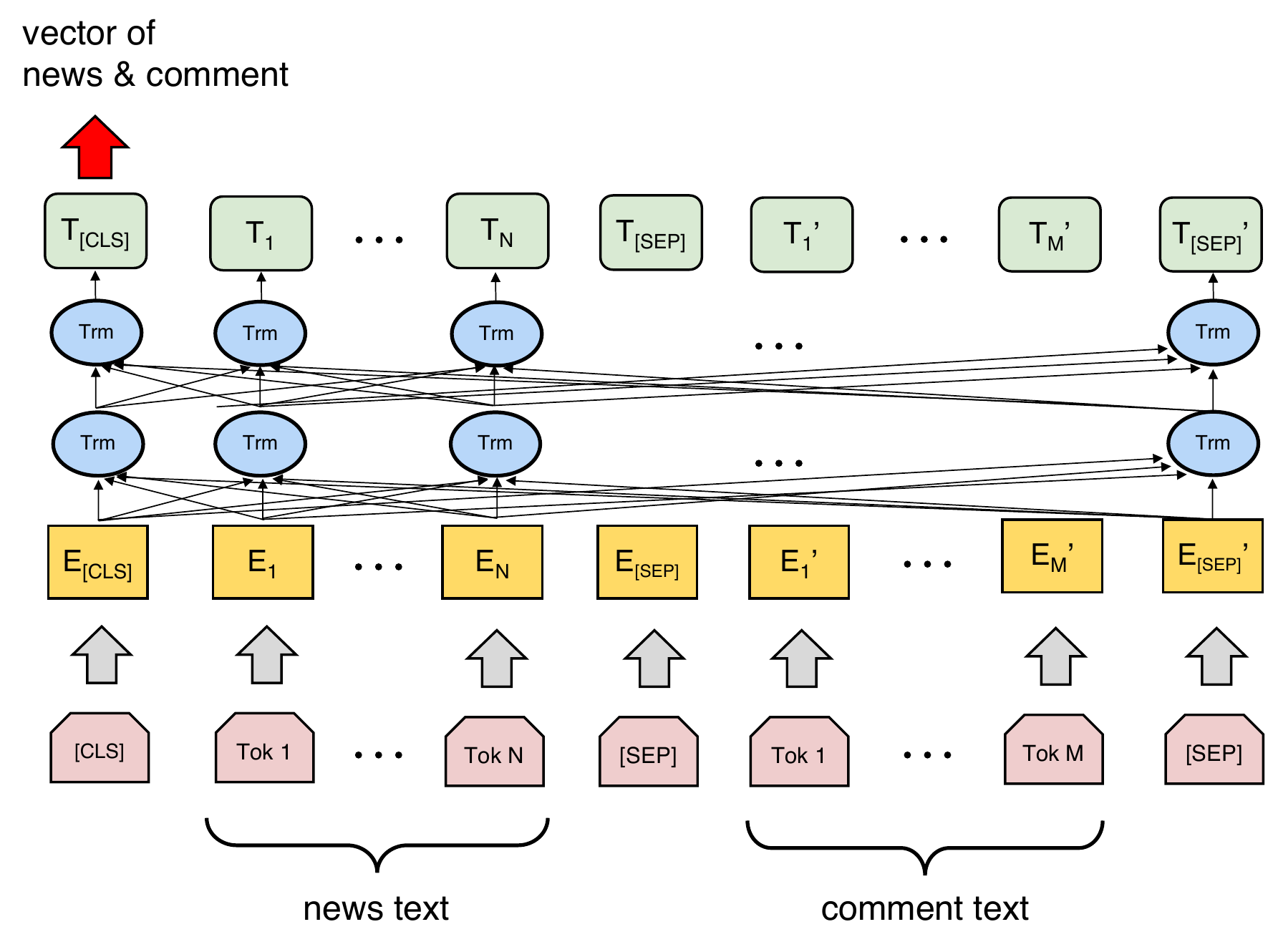}
\end{center}
\caption{BERT architecture}
\label{fig:BERT architecture}
\end{figure}

This section describes the vectorization of news and comments using BERT, which can transform input text into contextual vectors by considering word-to-word relationships and conduct various natural language processing tasks, such as sentence classification, with high accuracy. In this study, news and comment texts are combined in the following format, and the text divided into tokens is used as input to BERT:
\begin{displaymath}
\mbox{[CLS] news text [SEP] comment text [SEP]},
\end{displaymath}
where [CLS] is a special symbol that is added before each input example, and [SEP] is a special separator token. Using the self-attention mechanism in the transformer encoder\cite{NIPS2017_3f5ee243}, BERT outputs a vector that considers the relationship between words in the input data. Figure \ref{fig:BERT architecture} shows the architecture of BERT used in this study, where E is the input embedding, Trm is the transformer encoder, and T is the output from the last hidden layer of the transformer encoder. The output vector of the [CLS] token from the last hidden layer of the transformer encoder is used as the vector of news and comments for the method of predicting offensive news comments described in Chapters \ref{Personalization of Offensive News Comments Prediction} and \ref{Proposed Method}.

\subsection{Problems with Simple Offensive News Comment Prediction Model}
\label{Problems with Simple Offensive News Comments Prediction Model}
This section describes the problems with the simple prediction model stated in Section \ref{Simple Offensive News Comments Prediction Method}. This method uses the feedback of news comments that readers rated as ``offensive’’ in the past to personalize the prediction. However, providing a large amount of feedback is burdensome for readers. Therefore, it is desirable to use only a small amount of feedback data for prediction. However, if the amount of feedback data used for prediction is limited, the prediction results may be strongly influenced by the words and topics in the feedback. For example, we assume that only comments on political news are included in the feedback. In that case, correct predictions for comments on political news may be possible, although correct predictions for comments on sports news may be challenging. Similarly, it may be difficult to predict comments on the latest news that is not included in the feedback.

\section{Proposed Method}
\label{Proposed Method}
This section describes our proposed method for predicting offensive news comments that considers the commenter’s characteristics.

\subsection{Overview of the Proposed Method}
\label{Overview of the Proposed Method}
\begin{figure*}[t]
\begin{center}
\includegraphics[width=150mm]{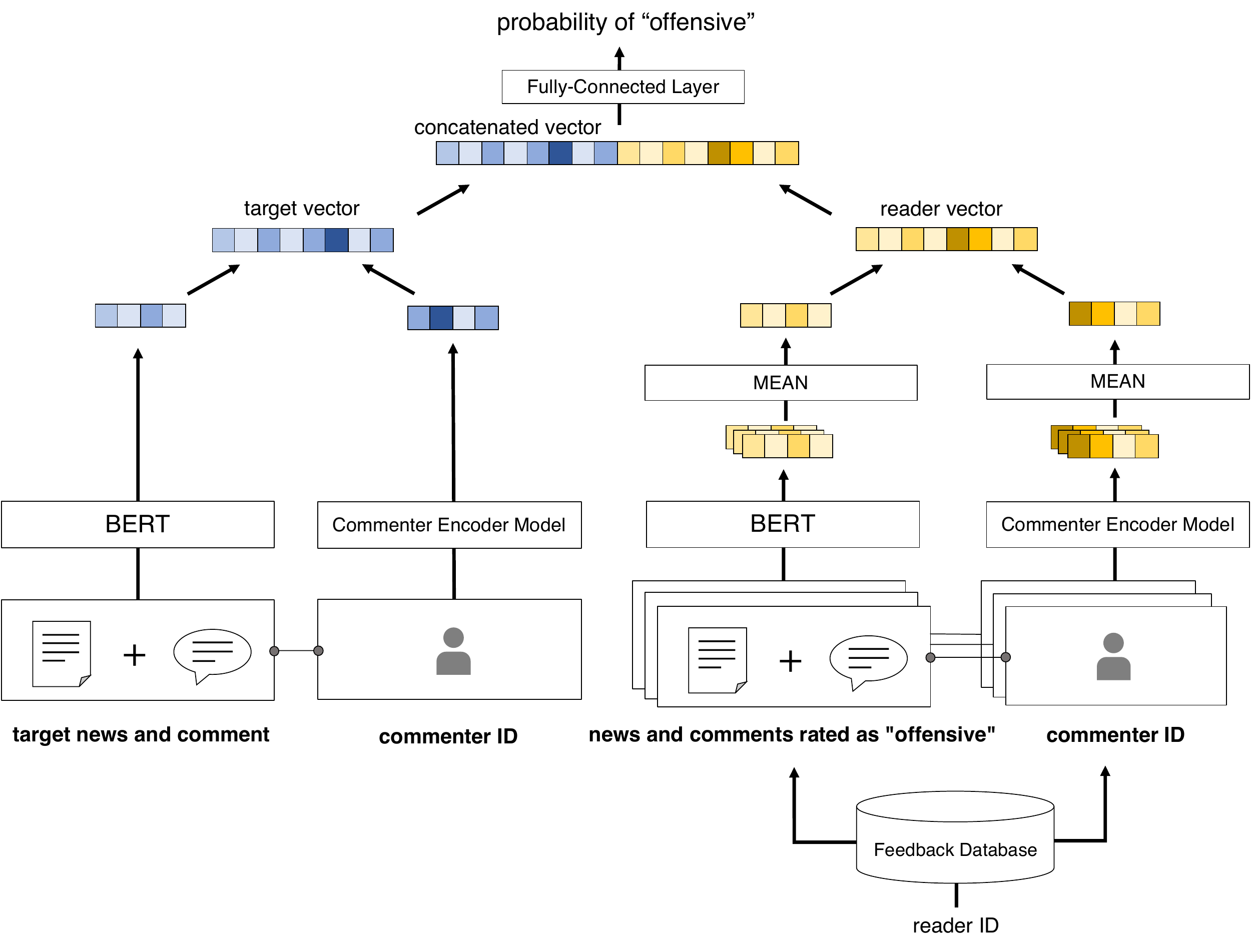}
\end{center}
\caption{Overview of the proposed method}
\label{fig:Overview of the proposed method}
\end{figure*}

Figure \ref{fig:Overview of the proposed method} shows an overview of the proposed method. The proposed method outputs the probability that the news comment to be predicted belongs to the ``offensive’’ label based on a vector that concatenates the target and reader vectors. In addition, the proposed method considers the characteristics of the commenters in the prediction. In Figure \ref{fig:Overview of the proposed method}, the characteristics of the commenters are used when generating the target and readers vector.

To personalize the predictions, we use feedback from news comments that readers have rated as ``offensive’’ in the past. The simple prediction model described in Section \ref{Simple Offensive News Comments Prediction Method} is expected to be strongly influenced by the words and topics in the feedback. The proposed method, a machine learning model using the features of commenters, considers the features of the user who posted the news comment that the reader rated as ``offensive’’ and the features of the user who posted the news comment to be predicted. This method is expected to result in predictions independent of the words and topics included in the feedback.

The aforementioned approach can be formalized as follows:
The proposed machine learning model that predicts the probability that user $r$ finds comment $c$ to be offensive is represented by the following equation:
\begin{equation}
y=f^{pro}(x;\theta^{pro}),
\end{equation}
where $x$ is a vector representation of the pair of reader $r$ and comment $c$, $y$ is a scalar value representing the probability value, and $\theta^{pro}$ is the set of parameters.
The training data set $D^{pro}$ for the proposed model is defined as the set of pairs $(x,y)$ of input $x$ and output $y$ as follows:
\begin{eqnarray}
D^{pro}=\{(x,y)\mid x=em^{pro}_c(c)\oplus em^{pro}_r(r), c\in C, r\in R, y\in \{1,0\} \}
\end{eqnarray}
where $x$ is the concatenation of the embedding vector of comment $c$ and the embedding vector of reader $r$.
In addition, $em^{pro}_c(c)$ and $em^{pro}_r(r)$ represent the proposed vectorization of comment $c$ and reader $r$, respectively.

\subsection{Generating Target Vectors for the Prediction}
\label{Generating Target Vectors of the Prediction}
To generate the target vectors, pairs of news and comments to be predicted are converted into vectors using BERT. Next, the users who posted the news comments to be predicted are vectorized by the commenter encoder model described in Section \ref{Commenter Encoder Model}. Finally, we generate the target vector by concatenating the vector of news and comments with the vector of the commenter.
The vectorization of the target comment $c$ in the proposed method can be defined as follows:
\begin{eqnarray}
em^{pro}_{c}(c)=em^{sim}_{c}(c)\oplus \mathit{enc}(\mathit{commenter}(c)),
\end{eqnarray}
where $\mathit{enc}(u)$ denotes the function to vectorize the commenter $u$ with the proposed commenter encoder model.

\subsection{Generating Reader Vectors}
\label{Generating the Reader Vectors}
To generate reader vectors, the reader ID is first entered. Then, based on the reader ID, we retrieve the texts of pairs of news and comments that the target reader has rated as ``offensive’’ in the past and user IDs of the commenters who posted the comments from the Feedback Database. Each output vector obtained by inputting the acquired news and comments to BERT is averaged. This produces a vector of news and comments that the reader has rated as ``offensive’’ in the past. Next, the users who posted comments that the reader rated as ``offensive’’ in the past are vectorized for each user using the commenter encoder model, and each vector is averaged. Finally, a reader vector is generated by concatenating the vectors of news and comments rated as ``offensive’’ and the vector of the user who posted the comment that was rated as ``offensive.’’
The proposed vectorization function of a reader $r$ is defined as follows:
\begin{eqnarray}
em^{pro}_r(r) = em^{sim}_{r}(r)\oplus \sum_{x \in \mathit{offensive}(r)}{\frac{ \mathit{enc}(\mathit{commenter}(x))}{\mid \mathit{offensive}(r)\mid}}
\end{eqnarray}

\section{Commenter Encoder Model}
\label{Commenter Encoder Model}
A machine learning model was proposed to vectorize the characteristics of commenters using the pairs of news and comment text; it has been proven that the vector adequately represents the characteristics of commenters\cite{nakahara2022}. The proposed method uses the same method to vectorize the features of commenters for prediction through a machine learning model that considers the features of commenters.

\subsection{Method for Vectorizing Characteristics of Commenters}
\label{Method for Vectorizing Characteristics of Commenters}
\begin{figure}[tb]
\begin{center}
\includegraphics[width=100mm]{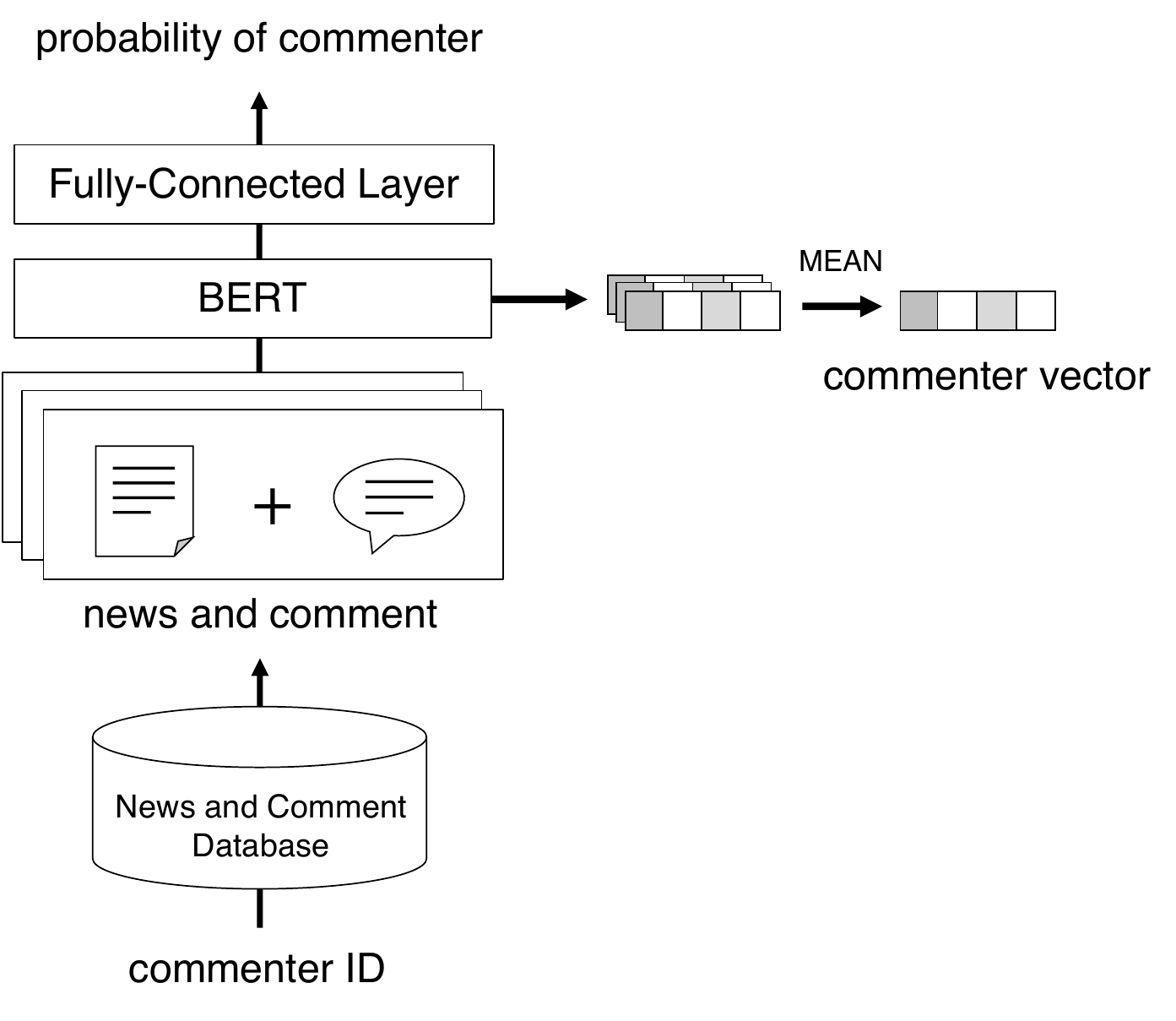}
\end{center}
\caption{Overview of the commenter encoder model}
\label{fig:Overview of the commenter encoder model}
\end{figure}

This section describes the commenter encoder model used to vectorize the characteristics of commenters. Figure \ref{fig:Overview of the commenter encoder model} shows an overview of this model. First, we construct a database that stores past news and comments along with the commenter IDs, which we refer to as the News and Comment Database. The commenter encoder model retrieves multiple texts of pairs of comments posted in the past by the commenter and the news from the News and Comment Database by entering the commenter ID. In this study, the number of retrieved cases was set to five. The retrieved text of pairs of news and comments is converted into a vector using BERT, and commenters are classified using the fully connected layer. By predicting the commenters, we can embed the characteristics of the commenters into the vectors in the hidden layer of the prediction model. In this method, the BERT layer in predicting commenters is used to generate a vector of commenters, and the average of the output of the BERT layer is used as the commenter vector for predicting offensive news comments.
When $N$ pairs of news and comments are obtained from a commenter $u$, let $\{d_{1}, d_{2}, \ldots, d_{N}\}$ be the text set of pairs of news and comments and $g(d_{k})$ be the output obtained from text $d_{k}$ by the BERT layer. The following formula defines the function to vectorize the commenter $u$.

\begin{equation}
enc(u) = \frac{1}{N}\sum_{k=1}^{N}g(d_{k})
\end{equation}

\subsection{Definition of the Training Dataset for the Commenter Encoder Model}
The commenter encoder model predicts the commenter from the input texts of news and comment pairs and embeds the commenter's features into the vector in the hidden layer. Therefore, to incorporate the commenter encoder model into the proposed method, we first train the commenter prediction in the commenter encoder model. 
Given a pair of news and comment $(nt,c)$, we consider predicting the commenter. Let P$(u \mid nt,c)$ denote the probability that the commenter is $u$. We assume that $\sum_{u \in U}P(u \mid nt,c) =1$. 
The proposed method uses a machine learning model that predicts P$(u \mid nt,c)$ to generate an embedding of the commenter. Let $(x,y)$ denote the training data used to train the prediction model, where $x$ is the input data, and $y$ is the ground truth for that data. We define the training dataset $D^{com}$ of the commenter encoder model as follows:
\begin{eqnarray}
D^{com}=\{(x,y) \mid x = BERT(c,news(c)), y=\mathit{commenter}(c), c \in C, y \in U\}
\end{eqnarray}


\section{Experiments}
\label{Experiments}

\subsection{Data and Settings}
\label{Data and Settings}

\subsubsection{Commenter Encoder Model}
\label{Evaluation：Commenter Encoder Model}

We constructed the News and Comment Database, which comprises news tweets posted by NHK News (@nhk\_news)\footnote{https://twitter.com/nhk\_news} from November 11, 2021 to March 31, 2022, users' replies to the news, and  user IDs of the users who replied to the news, which are referred to as news, comments, and commenter IDs, respectively. These were obtained using the Twitter API.

The commenter encoder model used to vectorize the characteristics of commenters accepts pairs of news and comments as input and classifies commenters. The features of the commenters are embedded into the vectors in the hidden layer of the prediction process. We trained the commenter encoder model on 515 users with more than 50 comments in the News and Comment Database for use in the proposed method. The BERT model was trained by fine-tuning a pre-trained BERT model\footnote{https://github.com/cl-tohoku/bert-japanese} using Wikipedia, which was developed by Inui Lab at Tohoku University. We randomly obtained 50 pairs of news and comments from users; we used 40 of each user's posts as training data, five as validation data, and the remaining five as test data. The input to the model for one datum is a single pair of news and comment, and the output is a 515-dimensional probability vector.

\subsubsection{Model for Offensive News Comment Prediction}
\label{Model for Offensive News Comments Prediction}

We labeled news comments by conducting a survey on the crowdsourcing service CrowdWorks. The total number of subjects was 250, divided into five groups of 50 each. Each group was presented with 400 news comments posted during the three months from April 1 to June 30, 2022, and asked to respond on a 5-point scale regarding whether they found each news comment offensive \{1: Strongly disagree, 2: Disagree, 3: Neither agree nor disagree, 4: Agree, 5: Strongly agree\}. The news and comments presented to the subjects were the same in each group. Labels for 400 news comments were obtained per subject by assigning two labels to news comments with responses of 4 or 5 as ``offensive’’ and others as ``not offensive.’’ In other words, the total number of labels obtained was 250×400. The total number of each label is listed in Table \ref{tab:Total number of each label obtained in the survey}.

\begin{table}[tb]
\caption{Total number of each label obtained in the survey}
\label{tab:Total number of each label obtained in the survey}
\centering
\begin{tabular}{ll}
\hline\hline
Label & Total\\ \hline
not offensive & 70,800 \\
offensive & 29,200 \\
\hline
\end{tabular}
\end{table}

We constructed the Feedback Database, which contains the pairs of news and comments rated as ``offensive’’ during the two weeks from April 1 to April 14, 2022, the commenter ID of the user who posted each comment, and the reader ID of the user who made the rating. However, the Feedback Database does not store news and comments rated as ``not offensive.’’ Therefore, the proposed method generates a reader vector by accepting the reader IDs as inputs and acquiring the pairs of news and comments rated as ``offensive’’ and the commenter IDs from the Feedback Database. Finally, we use the generated reader vectors to personalize the prediction.

The input training data are the pairs of news and comments to be predicted as well as the reader IDs, and the correct answer labels are the two labels of ``offensive’’ or ``not offensive’’ provided by the readers. The news comments from April 15 to June 9, 2022, from June 10 to June 20, 2022, and from June 21 to June 30, 2022, were used as the training data, validation data, and test data, respectively, in chronological order.

We trained the model to predict offensive news comments on 192 comment readers for whom we had obtained at least five feedbacks from the Feedback Database. For the BERT model for vectorizing news and comments, we used a pre-trained BERT model based on Japanese Wikipedia by Inui Lab at Tohoku University and conducted fine-tuning. The two BERT layers shown in Figure \ref{fig:Overview of the proposed method} are trained without sharing weights. Because the act of providing a large amount of feedback is burdensome for readers, it is desirable to use only a small amount of feedback data from readers for personalization. Therefore, we limited the personalization to five news comments that were rated as ``offensive’’ in the past when the reader ID was entered.

\subsubsection{Pre-processing}
\label{Pre-processing}
The following preprocessing was applied to the news texts used in this study:
\begin{itemize}
    \item Remove URLs
    \item Remove symbols
    \item Remove hashtags (e.g., \#nhk\_news)
\end{itemize}

In addition, the following preprocessing was applied to the comment texts in this study:
\begin{itemize}
    \item Remove URLs
    \item Remove symbols
    \item Remove emojis
    \item Remove mentions (@user ID)
\end{itemize}

\subsubsection{Description of Comparison Models}
\label{Description of Each Model}
To evaluate and compare the results of the offensive news comment prediction, we compared the prediction results of three different models in this experiment. The details of each model are as follows:

\begin{description}
    \item[Simple Prediction Model]\mbox{}\\
    As shown in Figure \ref{fig:Simple prediction model for offensive news comments}, we generate a reader vector using multiple feedbacks of news comments that readers have rated as ``offensive’’ in the past and personalize the prediction of offensive news comments.

    \item[Proposed Model]\mbox{}\\
    As shown in Figure \ref{fig:Overview of the proposed method}, we personalize the prediction of offensive news comments by incorporating a structure that utilizes the characteristics of commenters into a simple prediction model.

    \item[Model without Personalization]\mbox{}\\
    We predict offensive news comments by using only the news comment to be predicted. Because the reader vectors are not used for prediction, the prediction is not personalized.
\end{description}

\subsection{Results and Discussion}
\label{Results and Discussion}

\subsubsection{Comparison by the precision-recall (PR) curve}
\label{Comparison by the precision-recall (PR) curve}
\begin{figure}[tb]
\begin{center}
\includegraphics[width=120mm]{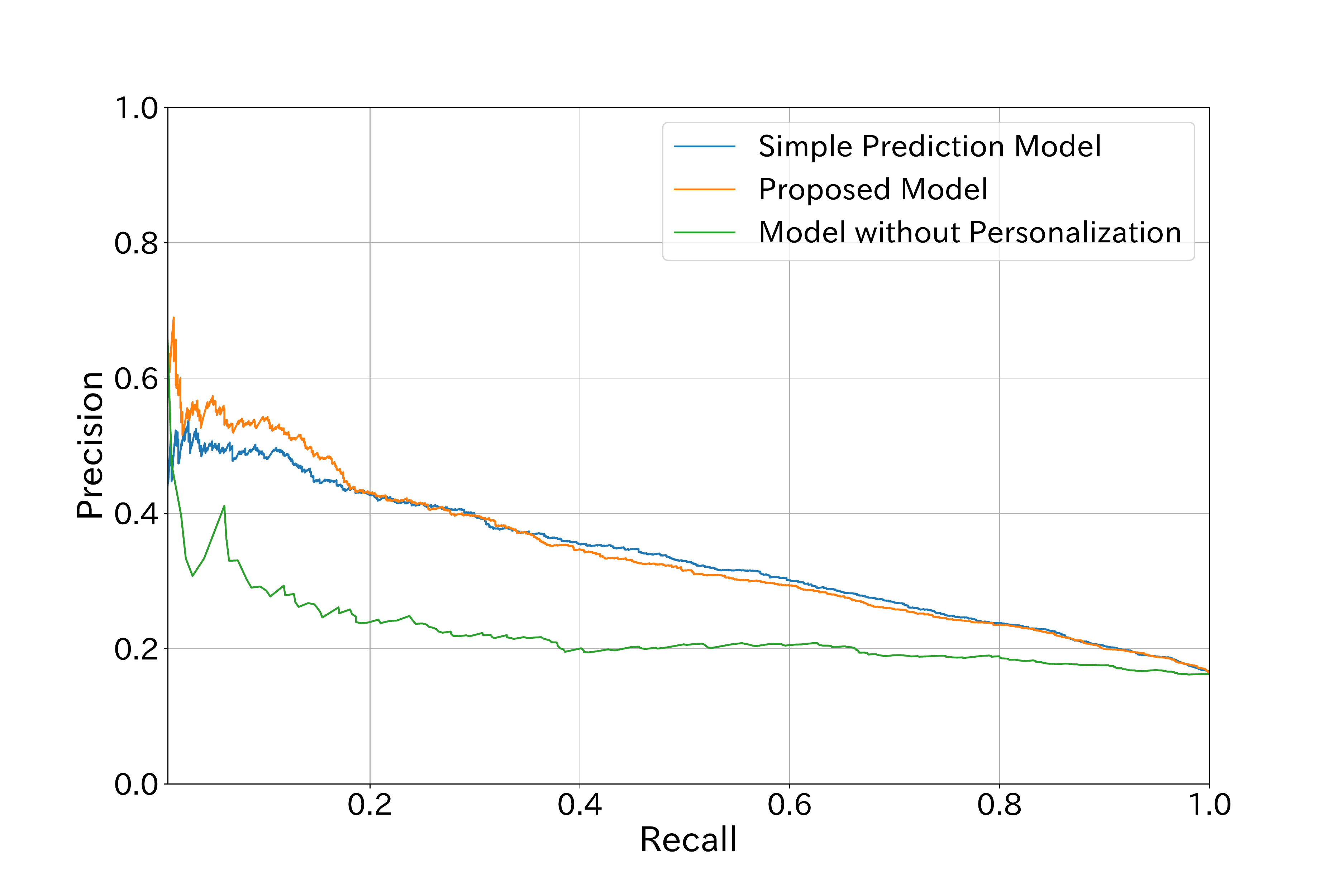}
\end{center}
\caption{PR curves}
\label{fig:PR curve}
\end{figure}

Figure \ref{fig:PR curve} shows a PR curve based on the prediction results of the three models on the test data. The PR curve is a plot of the relationship between precision and recall while changing the threshold for the prediction probability output by the prediction model. By comparing the precision for the same recall, the performance of the models can be compared without manually setting a threshold. PR curves are proved to be suitable for evaluating classifications with a biased distribution of labels, as demonstrated in this experiment. Table \ref{tab:AUC} summarizes the AUC values for the PR curves.

\begin{table}[tb]
\caption{AUCs of the PR curves}
\label{tab:AUC}
\centering
\begin{tabular}{ll}
\hline\hline
Model & AUC\\ \hline
Simple Prediction Model & 0.334 \\
Proposed Model & \bf0.339 \\
Model without Personalization & 0.223 \\
\hline
\end{tabular}
\end{table}

First, we examined the validity of the personalization of the predictions. Figure \ref{fig:PR curve} and Table \ref{tab:AUC} show that the prediction performance of the models with personalization (proposed model and simple prediction model) significantly exceeded that of the model without personalization. This result indicates that personalization of predictions is effective in predicting offensive comments.

Next, we compared the performance of the proposed model with that of the simple prediction model. Figure \ref{fig:PR curve} shows that the proposed model outperformed the simple prediction model in the recall range from 0.0 to 0.2. Moreover, the proposed model achieved lower or equal precision in the recall range from 0.2 to 1.0. These results indicate that the proposed method, which considers the characteristics of the commenter, is effective in situations with low recall for comments with high predictive probability and a low probability of false detection of offensive comments. Table \ref{tab:simple_threshold} lists the metrics for evaluating the simple prediction model when the threshold used to determine a news comment as ``offensive’’ or ``not offensive’’ is manually changed, and Table \ref{tab:proposed_threshold} shows the metrics for evaluating the proposed model.

\begin{table*}[tb]
  \begin{minipage}[tb]{.45\textwidth}
     \caption{Metrics for evaluating the simple prediction model at each threshold}
    \label{tab:simple_threshold}
    \begin{center}
        \begin{tabular}{ccccc}
        \hline\hline
        Threshold & Accuracy & Recall & Precision & F-measure \\
        \hline 
        0.9 & 0.838 & 0.003 & 0.333 & 0.005  \\
        0.8 & 0.838 & 0.052 & 0.500 & 0.095  \\
        0.7 & 0.833 & 0.146 & 0.448 & 0.220  \\
        0.6 & 0.817 & 0.282 & 0.404 & 0.332  \\
        0.5 & 0.779 & 0.431 & 0.351 & 0.387  \\
        0.4 & 0.707 & 0.607 & 0.300 & 0.402  \\
        0.3 & 0.582 & 0.767 & 0.246 & 0.373  \\
        0.2 & 0.416 & 0.899 & 0.204 & 0.333  \\
        0.1 & 0.232 & 0.986 & 0.172 & 0.293  \\
        \hline
      \end{tabular}
    \end{center}
  \end{minipage}
  \hfill
  \begin{minipage}[tb]{.45\textwidth}
     \caption{Metrics for evaluating the proposed model at each threshold}
    \label{tab:proposed_threshold}
    \begin{center}
    \begin{tabular}{ccccc}
    \hline\hline
    Threshold & Accuracy & Recall & Precision & F-measure \\
    \hline 
    0.9 & 0.839 & 0.012 & 0.667 & 0.023  \\
    0.8 & 0.839 & 0.064 & 0.532 & 0.114  \\
    0.7 & 0.837 & 0.145 & 0.491 & 0.224  \\
    0.6 & 0.824 & 0.225 & 0.419 & 0.293  \\
    0.5 & 0.804 & 0.329 & 0.380 & 0.352  \\
    0.4 & 0.763 & 0.450 & 0.330 & 0.381  \\
    0.3 & 0.698 & 0.607 & 0.292 & 0.394  \\
    0.2 & 0.576 & 0.763 & 0.242 & 0.368  \\
    0.1 & 0.357 & 0.938 & 0.193 & 0.321  \\
    \hline
      \end{tabular}
    \end{center}
  \end{minipage}
\end{table*}

\subsubsection{Comparison by Precision@k}
\label{Comparison by Precision@k}
The predicted probabilities for the test data output by the prediction model were placed in descending order. The percentage of correct labels when the top k data were determined as ``offensive’’ (Precision@k) was calculated for each reader. The model's performance was compared by averaging the Precision@k for each reader. Readers with fewer than k comments rated as ``offensive’’ in the test data were excluded. The values of Precision@k for k=1, 3, 5, and 10 are listed in Table \ref{tab:Precision@k}. Based on the total number of each label shown in Table \ref{tab:Total number of each label obtained in the survey}, the chance level of Precision@k is approximately 0.292 (29,200/100,000). Table \ref{tab:Precision@k} shows that the proposed model outperforms the simple prediction model in terms of prediction performance, with values greater than the chance level observed for both models. This result indicates that the prediction can be personalized even when the amount of readers' feedback data used in the prediction is small. Furthermore, the personalization of the proposed model, which considers the characteristics of the commenter, is less affected by the words and topics included in the feedback, which may have led to the improvement in prediction performance.

In addition, for small values of k, such as one or three, the proposed model outperforms the simple prediction model by a significant margin. As explained in Section \ref{Comparison by the precision-recall (PR) curve}, this is because the proposed model, which considers the characteristics of the commenter, is effective when targeting comments with a high predicted probability.

\begin{table}[tb]
\caption{Precision@k}
\label{tab:Precision@k}
\centering
\begin{tabular}{lcccc}
\hline\hline
Model  &  k=1 & k=3 & k=5 & k=10 \\
\hline 
Simple Prediction Model &  0.378 & 0.402 & 0.430 & 0.472  \\
Proposed Model & \bf{0.450} & \bf{0.484} &  \bf{0.445} & \bf{0.481} \\
\hline
\end{tabular}
\end{table}

\section{Conclusion}
\label{conclusion}
In this study, we attempted to personalize the prediction of offensive news comments based on a small amount of feedback from news comment readers who rated news comments as ``offensive’’ in the past. To generate predictions independent of the words and topics in the feedback, we proposed a machine learning model that considers the characteristics of the commenter. By analyzing the variations in the ratings of news comments, we found that variation exists in the ratings of comments among readers, indicating the importance of personalization in predicting offensive news comments. Furthermore, the experimental results showed that personalization of predictions is possible even when the amount of readers' feedback data used in the prediction is small. In particular, the proposed method, which considers the characteristics of the commenter, has a low probability of false detection of offensive comments.
In the future, we will attempt to improve the performance of the proposed method to accurately predict offensive news comments. The prediction model's current performance is insufficient for practical use, and the model needs to be improved. We are also investigating the effect of the amount of feedback data on the personalization of predictions by examining the change in prediction performance when the amount of readers' feedback data used in the prediction is varied.

\section*{Acknowledgments}
This research was partially supported by JSPS KAKENHI No.19H04219

\bibliographystyle{unsrt}  
\bibliography{nakahara}

\end{document}